\documentclass[conference]{IEEEtran}

\usepackage{times}
\usepackage{inconsolata}
\usepackage[numbers]{natbib}
\usepackage{multicol}
\usepackage[hidelinks,bookmarks=true,colorlinks=false]{hyperref}

\usepackage{amsmath, mathtools, amssymb, amsthm}
\usepackage{nicefrac}
\usepackage{mathrsfs}
\usepackage{siunitx}
\usepackage{caption}

\usepackage{algorithm}
\usepackage[noend]{algpseudocode}

\usepackage[capitalize,nameinlink]{cleveref}
\crefformat{equation}{(#2#1#3)}
\usepackage[acronyms, shortcuts, hyperfirst=false]{glossaries}
\glsdisablehyper

\usepackage{colortbl}

\usepackage{xcolor}

\newcommand{\R}{\mathbb{R}}

\newcommand{\parens}[1]{\left( #1 \right)}
\newcommand{\brackets}[1]{\left[ #1 \right]}
\newcommand{\braces}[1]{\left\{ #1 \right\}}
\newcommand{\norm}[1]{\left\lVert#1\right\rVert}

\newcommand*{\trans}{^{\mkern-1.5mu\mathsf{T}}}
\newcommand{\normal}{\mathcal{N}}
\newcommand{\partialderiv}[2]{\frac{\partial #1}{\partial #2}}

\newtheorem{rem}{Remark}
\newtheorem{assum}{Assumption}
\newtheorem{hypoth}{Hypothesis}

\crefname{line}{}{}

\newacronym{lstm}{LSTM}{long short-term memory}
\newacronym{fnn}{FNN}{feedforward neural network}
\newacronym{mppi}{MPPI}{Model Predictive Path Integral Control}
\newacronym{mpc}{MPC}{Model Predictive Control}
\newacronym{mse}{MSE}{mean squared error}
\newacronym{nll}{NLL}{negative log-likelihood}

\newcommand{\state}{x}
\newcommand{\ctrl}{u}
\newcommand{\inp}{y}
\newcommand{\pos}{p}
\newcommand{\vel}{v}
\newcommand{\bsr}{z}
\newcommand{\bsrmodel}{g_z}
\newcommand{\euler}{E}
\newcommand{\lstm}{\zeta}
\newcommand{\transform}{X}
\newcommand{\force}{F}
\newcommand{\cor}{D}
\newcommand{\cormodel}{g_{\cor}}
\newcommand{\params}{\phi}
\newcommand{\adaptparams}{\theta}
\newcommand{\paramparams}{\psi}
\newcommand{\paramdim}{n}
\newcommand{\forcemodel}{g_{\force}}
\newcommand{\nnin}{\mathbf{\eta}}

\newcommand{\controlseq}{\mathbf{\ctrl}}
\newcommand{\mppiwindow}{T}
\newcommand{\mppitrajs}{N}
\newcommand{\totalcost}{J}
\newcommand{\stagecost}{l}
\newcommand{\costtogo}{V}
\newcommand{\trackcost}{\mathtt{track}}
\newcommand{\rollovercost}{\mathtt{rollover}}
\newcommand{\othercost}{\mathtt{other}}

\newcommand{\polyfunc}{\mathcal{P}}
\newcommand{\rrlimit}{r_\mathtt{limit}}

\newcommand{\weightens}{\mathbf{W}}
\newcommand{\nnout}{\Phi}
\newcommand{\numin}{\paramdim_\mathtt{in}}
\newcommand{\numout}{\paramdim_\mathtt{out}}
\newcommand{\numenslayers}{\paramdim_w}

\newcommand{\numsteps}{h}
\newcommand{\paramcov}{P}
\newcommand{\proccov}{Q}
\newcommand{\meascov}{R}
\newcommand{\innovcov}{S}
\newcommand{\measmat}{H}
\newcommand{\procnoise}{w^{\adaptparams}}
\newcommand{\measnoise}{w^{\state}}
\newcommand{\selmat}{C}
\newcommand{\jac}{\mathcal{F}}
\newcommand{\startstep}{s}
\newcommand{\kadj}{\gamma}
\newcommand{\adj}{\varepsilon}
    
\newcommand{\numepochs}{N_E}
\newcommand{\batchsize}{N_B}
\newcommand{\dataset}{\mathcal{D}}
\newcommand{\adaptlen}{\tau}
\newcommand{\loss}{\mathcal{L}}

\newcommand{\dummy}{\xi}
\newcommand{\lr}{\alpha}
\newcommand{\decay}{\beta}

\newcommand{\confint}{\textsuperscript{*}}
\newcommand{\baseline}{\texttt{Baseline (no adaptation)}}
\newcommand{\ours}{\texttt{Meta-adaptation (ours)}}
\newcommand{\nometa}{\texttt{Adaptation}}
\newcommand{\leastsq}{\texttt{Sliding LSQ}}

\begin{document}

\title{Meta-Learning Online Dynamics Model Adaptation in Off-Road Autonomous Driving}


\author{\authorblockN{
    Jacob Levy\authorrefmark{1},
    Jason Gibson\authorrefmark{2},
    Bogdan Vlahov\authorrefmark{2}, 
    Erica Tevere\authorrefmark{3},
    Evangelos Theodorou\authorrefmark{2},\\
    David Fridovich-Keil\authorrefmark{1},
    Patrick Spieler\authorrefmark{3}
    }
    \authorblockA{\authorrefmark{1}University of Texas at Austin
    \authorrefmark{2}Georgia Institute of Technology}
    \authorrefmark{3}Jet Propulsion Laboratory, California Institute of Technology
}

\maketitle

\begin{abstract}
High-speed off-road autonomous driving presents unique challenges due to complex, evolving terrain characteristics and the difficulty of accurately modeling terrain-vehicle interactions.
While dynamics models used in model-based control can be learned from real-world data, they often struggle to generalize to unseen terrain, making real-time adaptation essential.
We propose a novel framework that combines a Kalman filter-based online adaptation scheme with meta-learned parameters to address these challenges.
Offline meta-learning optimizes the basis functions along which adaptation occurs, as well as the adaptation parameters, while online adaptation dynamically adjusts the onboard dynamics model in real time for model-based control. 
We validate our approach through extensive experiments, including real-world testing on a full-scale autonomous off-road vehicle, demonstrating that our method outperforms baseline approaches in prediction accuracy, performance, and safety metrics, particularly in safety-critical scenarios.
Our results underscore the effectiveness of meta-learned dynamics model adaptation, advancing the development of reliable autonomous systems capable of navigating diverse and unseen environments.
Video is available at: \url{https://youtu.be/cCKHHrDRQEA}
\end{abstract}

\IEEEpeerreviewmaketitle

\section{Introduction}

High-speed off-road autonomous driving presents a unique set of challenges where precise and reliable control is essential for traversing complex and unseen environments. 
These settings are characterized by diverse terrain types such as sand, snow, and dense vegetation, as well as varying terrain conditions like wetness, deformability, and roughness.
Such variability can significantly alter vehicle dynamics, introducing substantial uncertainties \cite{iagnemma2002terrain, angelova2007learning, rogers2012continuous, lupu2024magic}.
Just as a human driver adjusts their driving policy based on how the vehicle responds to the terrain, autonomous systems must adapt dynamically to maintain both performance and safety.

Autonomous systems are increasingly relied upon in scenarios where human intervention is impractical, too slow, or dangerous.
For instance, Mars rovers operate with onboard autonomy to identify obstacles and plan safe paths, as communication delays prevent real-time teleoperation \cite{verma2023autonomous, nesnas2021autonomy}. 
On Earth, autonomous vehicles hold significant potential for disaster response, where hazardous environments could render human operation unsafe \cite{kruijff2012rescue, kruijff2016deployment, lee2021strategic}.
Similarly, in mining and resource extraction, they can transport materials across rugged and hazardous terrains, improving operational efficiency and worker safety \cite{bathla2022autonomous}.
Safety is paramount in these scenarios, as entering dangerous zones or tipping the vehicle over could have severe consequences for mission success or operational integrity.
\begin{figure}[t]
    \centering
    \includegraphics[width=.48\textwidth]{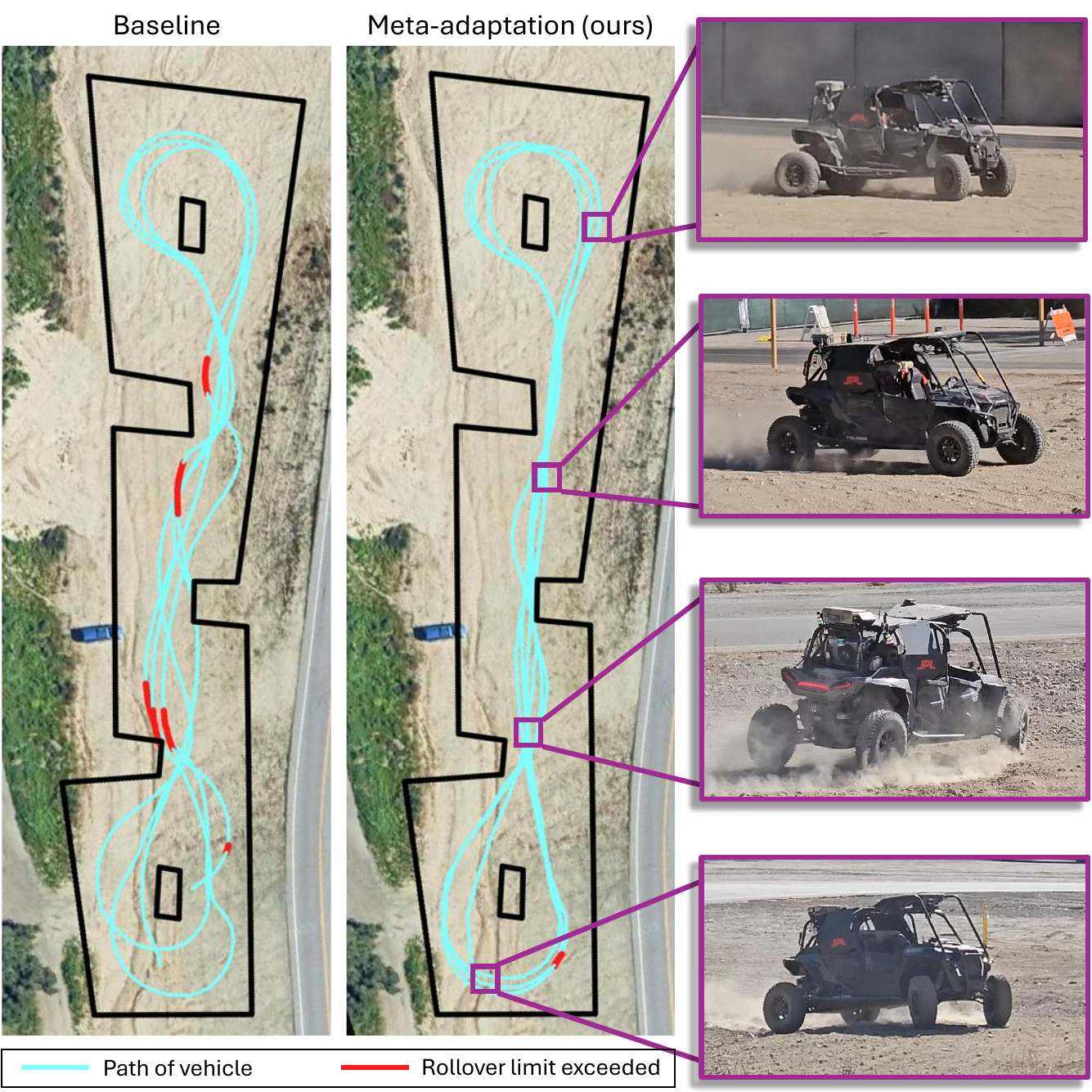}
    \caption{
    \footnotesize{
    Trajectories for a single 3-lap run, with insets displaying video stills. The baseline configuration shows erratic trajectories with frequent course boundary and rollover limit violations. In contrast, our adaptation configuration demonstrates more deliberate and compliant trajectories as the car learns the terrain dynamics in real time.
    }
    }
    \label{fig:front}
    \vspace{-1em}
\end{figure}

When system dynamics are known, model-based control techniques are widely used for the effective control of autonomous systems.
For instance, model predictive path integral (MPPI) control \cite{williams2016aggressive} is an algorithm that rolls out sampled control inputs on a dynamics model to identify optimal trajectories.
This method has demonstrated significant success in off-road autonomous driving applications \cite{cai2022risk, han2023model, lee2023learning, gibson2023multi, meng2023terrainnet, vlahov2024low}, but its performance hinges on the accuracy of the underlying dynamics model.
Accurately modeling interactions between vehicles and terrain remains a significant challenge, particularly in high-speed off-road driving, where changing terrain and operational conditions push the vehicle to its performance limits.
When the dynamics model fails to capture these interactions, it can lead to degraded performance and compromised safety.

\begin{figure*}[t]
    \centering
    \includegraphics[width=.99\textwidth]{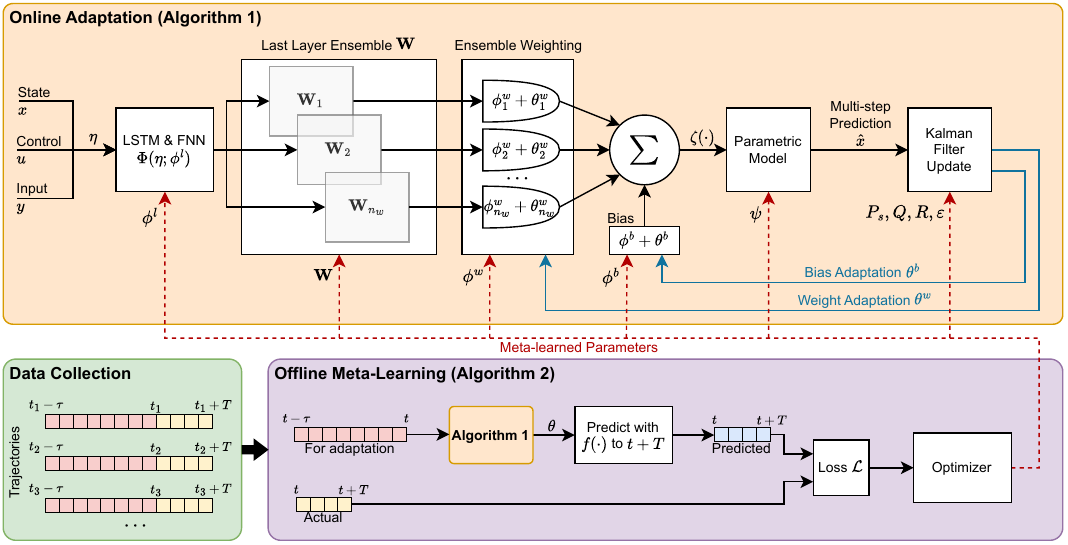}
    \caption{\footnotesize{
    Meta-learning online dynamics model adaptation. Online, a Kalman filter updates the linear combination of an ensemble of last-layer weights (\cref{alg:adapt}). Offline, trajectory segments are used to meta-learn model parameters, the last-layer ensemble, and the Kalman filter parameters (\cref{alg:meta}).
    }}
    \label{fig:method}
\end{figure*}

To address these limitations, hybrid approaches have emerged, integrating first-principles dynamics models with learned components to improve predictive accuracy \cite{djeumou2022neural, kim2022physics, djeumou2023autonomous, gibson2023multi, levylearning}.
While these models are effective under nominal conditions, they can struggle to generalize to previously unseen terrains or adapt to evolving dynamics during operation.
This limitation underscores the need for adaptive models capable of online adjustment.

We propose a novel framework (\cref{fig:method}) for meta-learned online dynamics model adaptation designed to address these challenges.
By leveraging meta-learning techniques, our method enables efficient and effective adaptation of the dynamics model to real-time sensor data.
This approach allows the system to adjust to changes in terrain and vehicle behavior dynamically, providing accurate predictions essential for safe and efficient control.
Our contributions include:
\begin{itemize}
    \item A meta-learning framework for \emph{offline} optimization of adaptation basis functions and parameters, along with dynamics model parameters.
    \item An efficient Kalman filter-based \emph{online} adaptation scheme to update model parameters in real time, addressing the challenges of noisy and delayed state measurements.
    \item Empirical validation of our method on a full-scale autonomous off-road vehicle and simulated environments, showcasing improved performance and safety metrics over baseline approaches.
\end{itemize}
By bridging the gap between predictive accuracy and adaptability, our framework establishes a foundation for safer and more effective high-speed off-road autonomy.

\section{Related Work}

\subsection{Off-road Autonomous Driving}
State-of-the-art approaches in off-road autonomous driving often rely on model-based control strategies, where accurate dynamics models are essential for planning and executing trajectories effectively \cite{han2023model, meng2023terrainnet, cai2022risk}.
To account for the complexity of terrain-vehicle interactions, some methods augment dynamics models with learned components derived from real-world driving data \cite{lee2023learning, gibson2023multi}.
However, these approaches typically assume static terrain conditions and lack mechanisms for adapting to changes in terrain type during operation.
To address the variability of terrain, some methods incorporate perceptual information into learned dynamics models to infer terrain properties, enabling more informed control \cite{cai2023probabilistic, pokhrel2024cahsor, gibson2024dynamics}.
While effective in many scenarios, these approaches lack the ability to adapt when the vehicle encounters previously unseen terrain.
Our work directly addresses this limitation by introducing a framework for online dynamics model adaptation.
By dynamically adjusting the dynamics model in real time, our method allows the vehicle to respond effectively to unseen terrain properties and evolving environmental conditions.

\subsection{System Identification / Adaptive Control}
In offline system identification, the parameters of a dynamics model are learned from pre-collected data, and the resulting model is then used for model-based control during online operation.
In autonomous driving, this approach has been applied to parametric models \cite{aghli2018online, badar2024vehicle}, hybrid models that combine learned and first-principles components \cite{cai2023probabilistic, gibson2023multi, dikici2024learning}, and even full black-box dynamics models \cite{wang2024pay}.
To enable adaptation during real-time operation, indirect adaptive control methods identify dynamics parameters online and update the dynamics model continuously, which is then used for control \cite{ioannou2012robust}.
Prior work has successfully demonstrated indirect adaptive control for autonomous driving applications \cite{khan2022fast, lua2023nonlinear, nagy2023ensemble, silaa2024indirect}.
However, these methods rely on manually designed basis functions and user-tuned learning rates, requiring significant user effort that could lead to suboptimal results.
In this work, we use meta-learning to automate both, enabling more efficient, accurate, and robust online parameter adaptation across diverse terrain and driving conditions. 

\subsection{Meta-learning}
Meta-learning, or ``learning to learn,'' is a paradigm that enables machine learning models to rapidly adapt to new tasks or environments by optimizing for efficient adaptation across task distributions \cite{finn2017model, harrison2020continuous, harrison2020meta}.
This often involves a two-phase process: an offline meta-training phase to prepare the model for adaptation and an online phase where the model adapts to a specific task or environment in real time.
In robotics, successful applications of meta-learning involve learning adaptation modules that infer latent environment representations \cite{visca2022deep, yu2020learning, shi2021meta, belkhale2021model, xiao2024safe}, optimizing controller parameters for rapid adaptation \cite{song2020rapidly, richards2023control}, and learning to adapt traversability maps to changing environments \cite{seo2023metaverse}. 
The closest line of work to ours involves using meta-learning to optimize a Kalman filter \cite{neuralfly} or a recursive least squares \cite{lupu2024magic} model adaptation scheme for model-based control. 
These methods rely on user-defined adaptation rates, whereas our approach utilizes meta-learning to dynamically determine optimal rates based on collected data.
Another related work, \cite{sanghvioccam}, uses meta-learning to optimize Kalman filter parameters that govern adaptation rates.
However, its prediction model targets performance metrics directly, bypassing system dynamics.
In contrast, our approach adapts a physics-informed dynamics model, incorporating physics-based principles into the adaptation process.

\section{Preliminaries}

\subsection{Vehicle Dynamics}
\label{sec:dyn}
We model the vehicle dynamics according to \cite{gibson2024multistep, gibson2024dynamics}; a brief overview is presented here. 
The dynamics are a discrete-time nonlinear system of the form
\begin{equation}
\label{eqn:dynamics}
    \state_{t+1} = f(\state_t, \ctrl_t, \inp_t),
\end{equation}
where $\ctrl_t \in \R^3$ is the control input consisting of the throttle, brake, and steering inputs; and $\inp_t \in \R^{14}$ is external sensor input containing roll, pitch, and  surface normal vectors at the four wheels.
The state of the vehicle $\state_t$ is fully observable and composed of $\state_t = [\pos_t, \vel_t, \bsr_t] \in \R^{10}$, where $\pos_t \in \R^3$ consists of the global position and yaw, $\vel_t \in \R^3$ consists of the body-frame forward and lateral velocities and yaw-rate, and $\bsr_t \in \R^4$ consists of the brake position, the steering column angle and velocity, and the engine speed.
Euler integration is used to propagate the dynamics with:
\begin{equation}
\label{eqn:integration}
    f(\state_t, \ctrl_t, \inp_t) = \euler_A R(\pos_t)\state_t + \euler_B
    \begin{bmatrix}
    \vel_t\trans & \dot{\vel_t}\trans & \dot{\bsr_t}\trans
    \end{bmatrix}\trans,
\end{equation}
where $\euler_A$ and $\euler_B$, are Euler integration matrices, $R(\cdot)$ is a rotation matrix which rotates velocities from the body frame to the global frame, and $\dot{\bsr_t}$ is computed using a pre-trained brake, steering, and engine model: $\dot{\bsr_t} = \bsrmodel(\state_t, \ctrl_t)$.

The acceleration $\dot{\vel_t}$ is predicted with a hybrid parametric and learned model:
\begin{equation}
    \label{eqn:accel}
    \dot{\vel_t} = M^{-1} \transform(\state_t) \brackets{\force_t + \lstm(\nnin_t; \params, \adaptparams_t)} + \cor_t,
\end{equation}
where $M^{-1}$ is the inverse of a mass matrix, $\transform(\cdot)$ transforms forces from the wheel frame to the body frame, $\lstm(\cdot)$ contains a \ac{lstm} network and a \ac{fnn} with learned parameters $\params \in \R^{\paramdim_\params}$ and adaptable parameters $\adaptparams_t \in \R^{\paramdim_\adaptparams}$ which predicts residuals to tire forces $\force_t$, and the input to the neural networks $\nnin_t$ is the following concatenation: $\nnin_t = [\state_t, \ctrl_t, \inp_t, \force_t]$.
The tire forces $\force_t$ and the Coriolis and drag accelerations $\cor_t$ are computed using parametric models $\force_t = \forcemodel(\state_t, \ctrl_t; \paramparams)$ and $\cor_t = \cormodel(\state_t; \paramparams)$ with parametric model parameters $\paramparams \in \R^{p_\paramparams}$.

\subsection{Control Architecture}
\label{sec:ctrl}
We adopt the control architecture in \cite{gibson2024multistep}, which is based on \ac{mppi} \cite{williams2016aggressive, vlahov2024low}, a sampling-based \ac{mpc} method.
Each time \ac{mppi} is called, it estimates the optimal control sequence $\controlseq^\ast_s = \{\ctrl_{s}, \ldots, \ctrl_{s+\mppiwindow}\}$ which minimizes the cost function:
\begin{equation}
\label{eqn:totalcost}
    \totalcost(\controlseq_s) =  \costtogo(\state_{s+1+\mppiwindow}) + \sum_{t=s}^{s+\mppiwindow} \brackets{\stagecost(\state_t, \ctrl_t, \inp_t) + \costtogo(\state_t)}
\end{equation}
by sampling $\mppitrajs$ sequences of controls $\{\controlseq^i_s\}_{i=1}^\mppitrajs$, rolling out each control sequence on dynamics \cref{eqn:dynamics} for $\mppiwindow$ steps, and computing the cost of each trajectory with \cref{eqn:totalcost}.
The cost-to-go $\costtogo(\cdot)$ comes from a higher-level state lattice planner that computes the optimal cost to the target waypoint using Dijkstra's algorithm \cite{dijkstra1959note}.
\begin{rem}
\Ac{mppi} depends on the dynamics model \cref{eqn:dynamics}; an accurate dynamics model is crucial for predicting accurate rollouts for the optimization process.
\end{rem}

The stage cost $\stagecost(\cdot)$ is the sum of the following components:
\begin{equation}
\begin{aligned}
    \stagecost(\state_t, \ctrl_t, \inp_t) &= \trackcost(\state_t) \\
    &+ \rollovercost(\state_t, \inp_t)\\
    &+ \othercost(\state_t, \ctrl_t, \inp_t).
\end{aligned}
\end{equation}  
The track cost $\trackcost(\cdot)$ is derived from a grid-based traversability map generated using perception data and semantic terrain classification.
Grid cells associated with more challenging terrain are assigned higher costs and traversal beyond user-defined boundaries incurs a significant cost.
The rollover cost $\rollovercost(\cdot)$ is computed as follows:
\begin{equation}
    \label{eqn:rollovercost}
    \rollovercost(\state, \inp) = \polyfunc_n\parens{\min\parens{\force^L, \force^R}; \rrlimit},
\end{equation}
where $\force^L, \force^R \in \brackets{0,1}$ are the mass-normalized normal forces acting on the left- and right-side tires, computed from the inclination of the ground, and the vehicle mass distribution, forward velocity, and steering angle; a value of $0$ and $1$ correspond to unloaded and fully loaded, respectively.
$\polyfunc_n(\cdot; \rrlimit)$ is a function which exhibits growth proportional to power $n$ of the input whenever the input falls below the limit $\rrlimit$.
Here, \cref{eqn:rollovercost} produces a high cost when the loading on the left or right wheels fall below the minimum loading threshold.
Other cost terms such as control effort and wheel slip are grouped into $\othercost(\cdot)$.
\section{Meta-learning Model Adaptation}
\label{sec:method}

In this section, we present our approach (\cref{fig:method}), detailing how we design the adaptable components of the dynamics model, the online scheme for real-time parameter updates, and the meta-learning framework for optimizing adaptation.

\subsection{Adaptable Parameters}
Selection of the adaptable parameters $\adaptparams$ is crucial in online dynamics model adaptation.
Adapting all parameters $\params$ of the \ac{lstm} and \ac{fnn} in real-time is impractical for online adaptation, as the sheer number of parameters would result in prohibitively slow updates with respect to the timescale of the changing dynamics involved with high-speed off-road driving.
On the other hand, adapting only the parameters of the parametric model $\paramparams$ would be insufficient because this model cannot fully capture complex terrain interactions that the learned model $\lstm(\cdot)$ is designed to address.
Adapting only the parameters of the last layer of such learned models has proven effective in prior work \cite{banerjee2020adaptive, neuralfly, harrison2020meta}; this approach yields adaptable parameters that are linear with respect to the dynamics, enabling faster adaptation.

To achieve adaptation fast enough for high-speed off-road driving, we further streamline adaptation by reducing the number of adapted parameters $\paramdim_\adaptparams$, focusing on only adapting the linear combination of an ensemble of last-layer weights (\cref{fig:method}).
We represent the ensemble of last-layer weights in the \ac{fnn} as a tensor $\weightens \in \R^{\numenslayers \times \numout \times \numin}$, where $\numenslayers$ denotes the ensemble size, and $\numin$ and $\numout$ are the input and output feature dimensions of the last layer, respectively.
The weight matrices of the ensemble are stacked along the first dimension to form this tensor. 
Now, we express how the adapted parameters appear in the learned model with:
\begin{equation}
    \lstm(\nnin; \params, \adaptparams) = \parens{\params^w + \adaptparams^w}\trans \weightens \nnout(\nnin; \params^{l}) + \params^b + \adaptparams^b,
\end{equation}
where $\nnout(\cdot) \in \R^{\numin}$ is the output of the second to last layer of the \ac{fnn}, the learned parameters consist of $\params = [\params^{l}, \params^w, \params^b, \weightens]$, and the adaptable parameters consist of $\adaptparams = [\adaptparams^w, \adaptparams^b]$.
The learned ensemble weighting $\params^w \in \R^{\numenslayers}$ is adapted by $\adaptparams^w$ and the last layer bias $\params^b \in \R^{\numout}$ is adapted by $\adaptparams^b$.

\begin{rem}
    \label{rem:linear}
    The dynamics \cref{eqn:dynamics} are a linear function of the adaptable parameters $\adaptparams$.
\end{rem}
\begin{rem}
    The tensor $\weightens$ is a set of basis functions, with their weighting dynamically adjusted through adaptation of $\adaptparams^w$.
\end{rem}

\subsection{Online Adaptation}
Here, we present the online adaptation scheme (\cref{alg:adapt}) implemented onboard the vehicle for dynamically updating the adaptable parameters $\adaptparams$.
We employ a Kalman filter to quickly adapt the dynamics model to incoming measurements since the adaptable parameters are linear with respect to the dynamics (\cref{rem:linear}).
Propagation of the dynamics \cref{eqn:dynamics} occurs at a very quick timescale; we run the Kalman filter at a slower rate to improve parameter adaptation by adapting parameters every $\numsteps$ time steps.
This allows us to mitigate the impact of noisy and delayed state measurements, account for the minimal changes in integrated quantities over single time steps, and capture longer-term trends in parameter behavior.
To form the Kalman filter, we assume the following:
\begin{assum}
The adaptable parameters evolve according to a random walk: $\adaptparams_{t+\numsteps} = \adaptparams_{t} + \procnoise_t$ with
 $\procnoise_t \sim \normal(0, \proccov)$.
\end{assum}
\begin{assum}
Explicitly denoting the dependence on $\adaptparams$, we assume the system dynamics have additive Gaussian noise: $\state_{t+1} = f(\state_t, \ctrl_t, \inp_t; \adaptparams_t) + \measnoise_t$, where $\measnoise_t \sim \normal(0, \meascov)$.
\end{assum}
\begin{algorithm}[t]
\caption{Online Adaptation}
\label{alg:adapt}
\begin{algorithmic}[1]
\State \textbf{Input:} starting step $\startstep$, initial parameter covariance $\paramcov_\startstep$
\State \textbf{initialize} $t \gets \startstep, \adaptparams_0 \gets 0$
\While{running}
    \State $\hat{\state}_{t+1:t+\numsteps} \gets$ Propagate dynamics \cref{eqn:dynamics} $\numsteps$ steps with $\adaptparams_t$
    \State $\measmat_{t+\numsteps} \gets$ Compute multi-step Jacobian with \cref{eqn:jac}
    \State $\bar{\paramcov}_{t+\numsteps} \gets \paramcov_{t} + \proccov$ \label{line:kstart}
    \State $\innovcov_{t+\numsteps} = \selmat\parens{\measmat_{t+\numsteps} \bar{\paramcov}_{t+\numsteps} \measmat_{t+\numsteps}\trans + \meascov}\selmat\trans$
    \State $K_{t+\numsteps} \gets \bar{\paramcov}_{t+\numsteps}\measmat_{t+\numsteps}\trans\selmat\trans\innovcov_{t+\numsteps}^{-1}$
    \State $\adaptparams_{t+\numsteps} \gets \adaptparams_{t} + \kadj_t K_{t+\numsteps}\selmat\parens{\state_{t+\numsteps} - \hat{\state}_{t+\numsteps}}$ \label{line:scale}
    \State $\paramcov_{t+\numsteps} \gets \bar{\paramcov}_{t+\numsteps} + K_{t+\numsteps} \selmat\measmat_{t+\numsteps} \bar{\paramcov}_{t+\numsteps}$ \label{line:kend}
    \State $t \gets t + \numsteps$
\EndWhile
\end{algorithmic}
\end{algorithm}
With the notation $a_{t:t+\numsteps} \triangleq \braces{a_{t}, a_{t+1}, \ldots, a_{t+\numsteps}}$, where $a$ is any variable, we perform an $\numsteps$-step dynamics propagation to obtain predicted states $\hat{\state}_{t+1:t+\numsteps}$.
Since we propagate the dynamics over multiple steps, we compute the multi-step Jacobian $\measmat_{t+\numsteps} = \nicefrac{\partial \hat{\state}_{t+\numsteps}}{\partial \adaptparams_t}$ with the recursion:
\begin{equation}
\label{eqn:jac}
\begin{aligned}
    \measmat_{t+i} &=  \jac^\state_{t+i} \measmat_{t+i-1} + \jac^\adaptparams_{t+i} \\
    \measmat_{t} &= \jac^\adaptparams_{t},
\end{aligned}
\end{equation}
where:
\begin{equation}
\begin{aligned}
    \jac^\state_{t+i} &= \partialderiv{}{\state_{t+i}} f(\hat{\state}_{t+i}, \ctrl_{t+i}, \inp_{t+i}; \adaptparams_t) \text{ and} \\
    \jac^\adaptparams_{t+i} &= \partialderiv{}{\adaptparams_t} f(\hat{\state}_{t+i}, \ctrl_{t+i}, \inp_{t+i}; \adaptparams_t).
\end{aligned}
\end{equation}
By performing a multi-step propagation, we include the dynamics Jacobian $\jac^{\state}$ in the multi-step Jacobian computation \cref{eqn:jac}.
This Jacobian captures how the current state influences future states, and its recursive incorporation  over multiple steps provides a richer, long-term understanding of how the adapted parameters $\adaptparams$ influence the dynamics, improving the accuracy of adaptation.
We make the following assumption to simplify computation of the multi-step Jacobian:
\begin{assum}
    Changes in the learned model output with respect to changes in the state are negligible, i.e., $\partial \lstm / \partial \state \approx 0.$
\end{assum}
We use the measurement selection matrix $\selmat \in \R^{3 \times 10}$ in \cref{alg:adapt} to select only the velocity measurements $\vel$ for parameter updates.
The position measurements $\pos$ are disregarded because the changes in position within a single Kalman filter update are negligible compared to the measurement noise.
Additionally, $\bsr$ is excluded: propagation of the brake and steering states does not depend on the adaptable parameters $\adaptparams$, and the engine state is omitted to simplify the multi-step Jacobian.
With $\hat{\state}_{t+1:t+\numsteps}$ and $\measmat_{t+i}$ available, we perform a Kalman update with lines \cref{line:kstart} to \cref{line:kend} of \cref{alg:adapt}.
To prevent unnecessary parameter updates while the vehicle is moving very slowly, we scale the update in line \cref{line:scale} with:
\begin{equation}
    \kadj_t = \frac{\norm{\vel_t}^2_2}{\norm{\vel_t}^2_2 + \adj},
\end{equation}
where $\adj > 0$ adjusts the scaling intensity.
The updated parameters are then used with the controller (\cref{sec:ctrl}) for the next $\numsteps$ time steps.

\subsection{Offline Meta-learning}
With the adaptation framework established, our goal is to optimize the basis functions $\weightens$ and the Kalman filter parameters $\paramcov_\startstep$, $\proccov$, $\meascov$, and $\adj$ \emph{offline} to maximize the effectiveness of online adaptation.
This is achieved through a two-phase process: first, a data collection phase to gather diverse and representative trajectories, followed by a training phase to refine the parameterization and basis functions.

\subsubsection{Data Collection}
We begin by collecting autonomous driving data on a diverse set of terrains and conditions.
Data are collected in discrete \emph{runs}, with each run defined as a continuous period during which the robot is actively operating.
Each run is subsequently partitioned into potentially overlapping trajectories, stored in dataset $\dataset$, where each trajectory is represented as $\braces{\state_{t-\adaptlen:t+\mppiwindow}, \ctrl_{t-\adaptlen:t+\mppiwindow}, \inp_{t-\adaptlen:t+\mppiwindow}}$, encapsulating the state, control inputs, and external inputs over a period of $\adaptlen + \mppiwindow$ time steps.
Here, each trajectory will be associated to a unique reference time $t$, $\adaptlen$ is the adaptation length, and $\mppiwindow$ is the \ac{mppi} prediction length (\cref{sec:ctrl}).
Each trajectory is short relative to the overall duration of the runs---on the order of seconds. 
Consequently, individual trajectories are generally limited to a single terrain type and a specific terrain condition, providing focused and localized data.

\begin{algorithm}[t]
\caption{Offline Meta-Learning}
\label{alg:meta}
\begin{algorithmic}[1]
\State \textbf{Input:} Dataset $\dataset$
\State \textbf{Output:} Meta-learned parameters $\params, \paramparams, \paramcov_\startstep, \proccov, \meascov, \adj$ 
\For{$\numepochs$ epochs}
    \While{$\dataset \neq \emptyset$} \Comment{iterate over entire dataset}
        \State Sample $\batchsize$ trajectories from $\dataset$ w/o replacement
        \For{each trajectory in batch}
            \State $\adaptparams_t \gets$ Run \cref{alg:adapt} from $t-\adaptlen$ to $t$
            \State Rollout dynamics \cref{eqn:dynamics} from $t$ to 
            $t+\mppiwindow$ with $\adaptparams_t$
            \State Compute trajectory loss $\loss_i$ from $t$ to $t+\mppiwindow$
        \EndFor
        \For{each $\dummy \in \braces{\params, \paramparams, \paramcov_\startstep, \proccov, \meascov}$}
            \State $\dummy \gets \dummy - \lr \nabla_{\dummy} \sum_{i=1}^{\batchsize} \loss_i$ \Comment{optimizer step}
        \EndFor
    \EndWhile
\EndFor
\end{algorithmic}
\end{algorithm}

\subsubsection{Training}
We perform offline training (\cref{alg:meta}) with gradient-based meta-learning \cite{finn2017model}, which consists of an inner $\adaptparams$ adaptation phase and an outer loop which optimizes $\params$, $\paramparams$ $\paramcov_\startstep$, $\proccov$, $\meascov$, and $\adj$ (the tensor $\weightens$ is reshaped as a vector and contained within learned parameters $\params$).
Given that each trajectory typically occurs within a single terrain type and condition, we make the following assumption:
\begin{assum}
    \label{assum:theta}
    The true value of $\adaptparams$ is slowly time-varying, i.e. approximately constant, for each trajectory.
\end{assum}
For each sampled trajectory, with \cref{assum:theta}, we can identify the adaptable parameters at the $t$-th time step, $\adaptparams_t$, by running \cref{alg:adapt} for a $\adaptlen$-step adaptation period from $t-\adaptlen$ to $t$.
\begin{rem}
\label{rem:decay}
During offline meta-learning, we introduce a gradual decay to the adapted parameters with $\theta_t \gets \decay \theta_t, \decay \in (0, 1)$.
This slight modification to \cref{alg:adapt} enhances training stability by preventing $\adaptparams$ from becoming excessively large, particularly in the early stages of training when the Kalman filter parameters remain uncalibrated.
\end{rem}
Once adaptation is complete, $\adaptparams_t$ and the dynamics model \cref{eqn:dynamics} are used to generate a $\mppiwindow$-step dynamics prediction ($\hat{\state}_{t+1:t+1+\mppiwindow}$).
The $i$-th predicted trajectory is then evaluated against the corresponding ground truth trajectory using the multi-step loss function $\loss_i$\footnote{For simplicity, we present the \ac{mse} loss, however, in practice, we use the \ac{nll} loss from \cite{gibson2024multistep}}:
\begin{equation}
    \loss_i = \frac{1}{T} \sum_{j=0}^\mppiwindow \norm{\hat{\state}_{t+1+j} - \state_{t+1+j}}^2_2,
\end{equation}
where $t$ is the reference time associated with the $i$-th trajectory.
Importantly, $\loss_i$ is indirectly a function of $\params, \paramparams, \paramcov_\startstep, \proccov$, $\meascov$, and $\adj$, which enables meta-learning of these parameters through backpropagation.
During this process, gradients are propagated through the entire procedure of \cref{alg:adapt}, allowing the meta-learning framework to optimize both the adaptation dynamics and the underlying Kalman filter parameters effectively.
\begin{rem}
    Since $\weightens$ is optimized through the meta-learning process, meta-learning effectively determines the ``directions'' in the parameter space along which the Kalman filter can adapt.
    In other words, it specifies the subset of parameters that the Kalman filter targets for adaptation, guiding the adaptation process to focus on the most relevant and impactful aspects of the system's dynamics.
\end{rem}
\begin{figure}[t]
\centering
    \includegraphics[width=.49\textwidth]{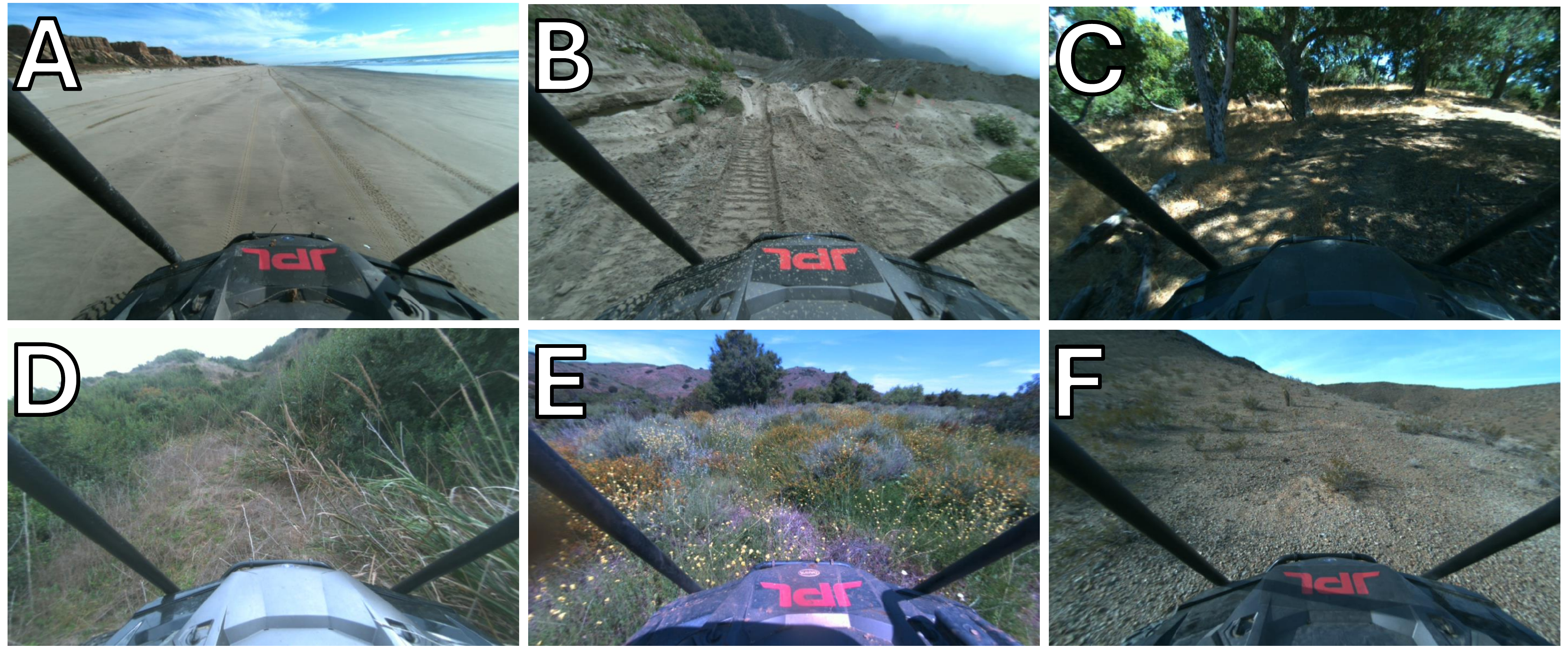}
\caption{\footnotesize{Forward facing camera stills from the dataset highlighting a diverse range of terrains: A) flat sandy beach with a mixture of packed wet sand and loose dry sand; B) wet dense mud that forms deep ruts; C) dirt trails with low dry grass that weave through dense trees; D) mixed vegetation including dry, dense vehicle-height grass; E) dense overgrown mixed vegetation ranging in crushability; and F) loose gravel, uneven ground, and steep slopes.}}
\label{fig:varied_terrains}
\end{figure}
\begin{table}[h]
    \centering
    \caption{\footnotesize{Dataset statistics.}}
    \begin{tabular}{lccccccc}
    \hline
        ~ & Mean & Std & Min & 5th \% & Median & 95th \% & Max  \\
        \hline
        Fwd. vel. (m/s) & 4.55 & 2.91 & -2.79 & -0.01 & 4.56 & 9.62 & 15.25  \\ 
        Lat. vel. (m/s) & 0.02 & 0.20 & -1.31 & -0.32 & 0.00 & 0.37 & 1.34  \\ 
        Yaw rate (rad/s) & 0.00 & 0.19 & -1.32 & -0.32 & 0.00 & 0.31 & 1.40  \\ 
        Pitch (deg) & -0.4 & 4.9 & -24.2 & -8.0 & -0.4 & 7.9 & 31.3  \\ 
        Roll (deg) & 0.2 & 4.8 & -28.7 & -7.8 & 0.2 & 8.7 & 28.8  \\ 
        \hline
    \end{tabular}
    \label{table:dataset}
\end{table}
\begin{table*}[tb]
    \centering
    \caption{
    \footnotesize{
    Real-world validation results.
    The means and standard deviations over 4 runs of each configuration are displayed.
    }
    }
    \begin{tabular}{lccccccccc}
    \hline
        ~ & Completion & Average & Prediction & \multicolumn{2}{c}{\# times crossed limit} & \multicolumn{2}{c}{Time exceeds limit, \si{\second}} & \multicolumn{2}{c}{Cost ($\times 10^4$)} \\ 
        ~ & time, \si{\second} & speed, \si{\meter\per\second} & Error, \si{\meter} & Track & Rollover & Track & Rollover & Track & Rollover \\
        \hline
        \baseline & 154.6 $\pm$ 16.9 & 5.06 $\pm$ 0.58 & 4.88 $\pm$ 0.47 & 8.0 $\pm$ 1.8 & 13 $\pm$ 5.4 & 5.32 $\pm$ 1.81 & 6.64 $\pm$ 2.09 & 40.8 & 20.2 \\ 
        \ours & \textbf{130.9 $\pm$ 7.8} & \textbf{5.84 $\pm$ 0.33} & \textbf{3.10 $\pm$ 0.18} & \textbf{3.3 $\pm$ 2.1} & \textbf{3.8 $\pm$ 1.7} & \textbf{0.57 $\pm$ 0.53} & \textbf{0.85 $\pm$ 0.59} & \textbf{1.95} & \textbf{0.36} \\ \hline
    \end{tabular}
    \label{tab:hardware}
\end{table*}
\begin{figure*}[tb]
    \centering
    \includegraphics[width=.99\textwidth]{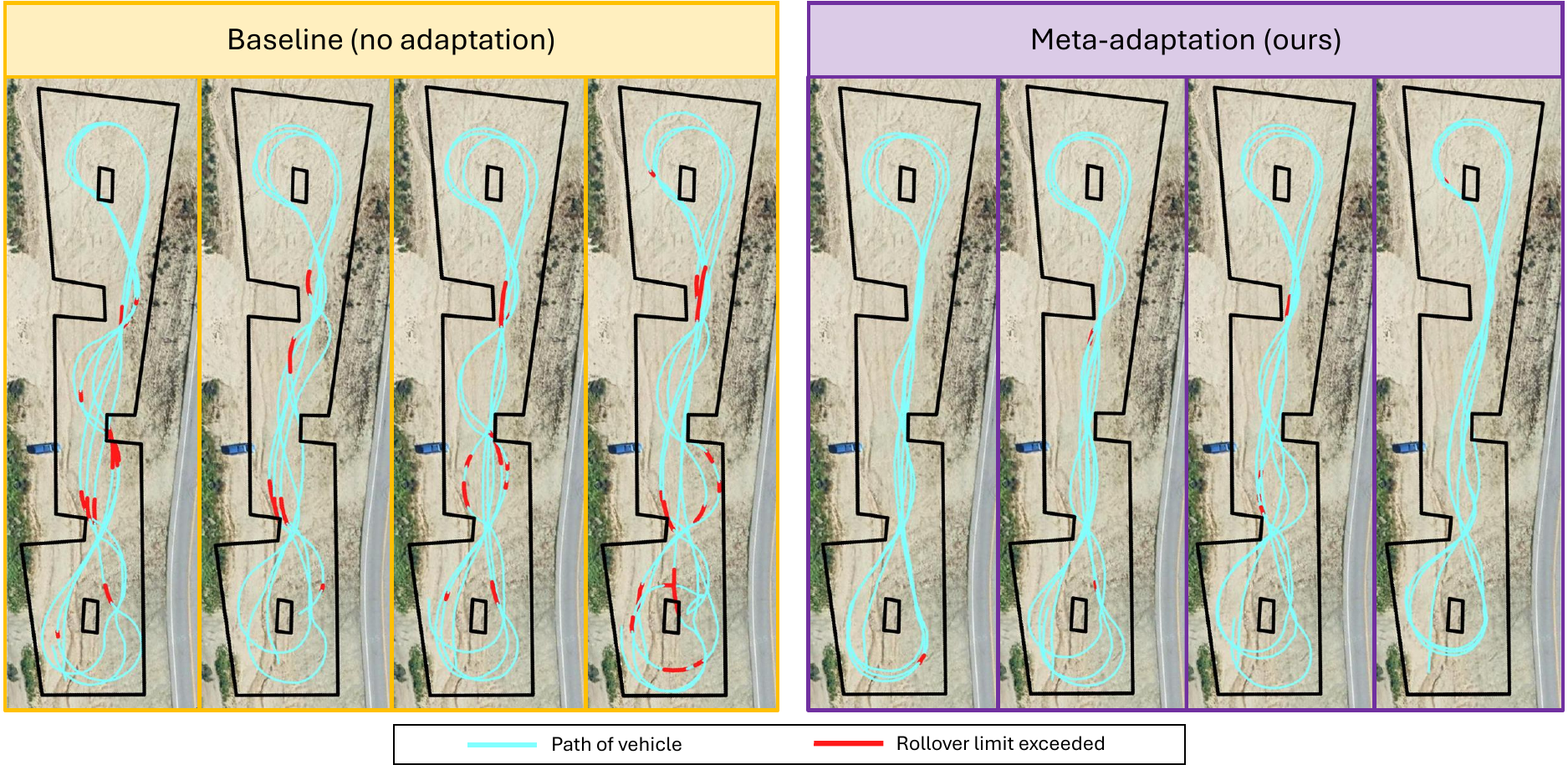}
    \caption{\footnotesize{
    Trajectories for all 3-lap real-world runs. The baseline configuration exhibits erratic motion, frequently violating course boundaries and rollover limits. In contrast, our adaptation configuration produces more deliberate and compliant trajectories, as the vehicle learns the terrain dynamics in real time.
    }}
    \label{fig:allruns}
\end{figure*}
\begin{rem}
    During the meta-learning process, the covariance matrices $\proccov$ and $\meascov$ are explicitly learned.
    These matrices play a critical role in determining the Kalman filter's time constant, as $\proccov$  reflects the expected variability of the adaptable parameters, and $\meascov$ encapsulates the process and measurement noise in the dynamics model and state observations. 
    Together, $\proccov$ and $\meascov$ govern how quickly the Kalman filter adapts $\adaptparams$ to changes in the environment, directly influencing the responsiveness and stability of the adaptation process.
\end{rem}
\begin{figure}[tb]
    \centering
    \includegraphics[width=.48\textwidth]{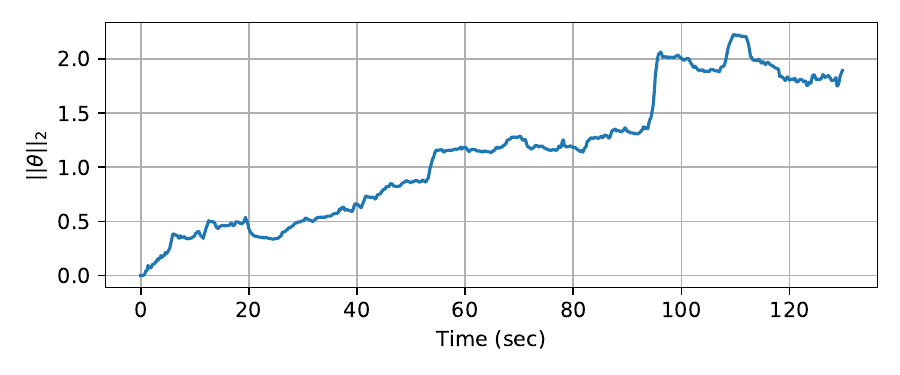}
    \vspace{-0.75em}
    \caption{\footnotesize{
    Norm of the adapted parameters during one of the real-world runs.
    }}
    \label{fig:thetanorm}
\end{figure}

\section{Experiments}
We validate our approach through both real-world and simulated experiments to investigate the following hypotheses:
\begin{hypoth}
\label{hypoth:pred}
Meta-learned model adaptation reduces the dynamics model's prediction error during online operation.
\end{hypoth}
\begin{hypoth}
\label{hypoth:perf}
Effective model adaptation enhances closed-loop behavior, improving both stability and safety.
\end{hypoth}

\subsection{Real-World Validation}

\subsubsection{Experimental Setup}
\label{sec:realexp}
We conducted real-world validation of our approach on a full-scale autonomous off-road vehicle: a modified Polaris RZR S4 1000 Turbo equipped with a 1.0\si{\liter} twin-cylinder engine, pictured in \cref{fig:front}.
The vehicle is outfitted with an extensive suite of onboard computation and sensors, including 2 LiDARs, 4 stereo/RGB cameras, IMUs, GPS, and wheel encoders.
State estimation is achieved using a localization module that integrates LiDAR and IMU data through a GTSAM-based Factor Graph Optimization (FGO) framework \cite{fakoorian2022rose, nubert2022graph}.
A 3D voxel map is constructed by fusing geometric and semantic data to generate traversability and elevation maps, providing track cost and ground slope \cite{atha2024few}.
The vehicle is controlled via a power-assisted steering actuator, a brake pressure pump, and electronic throttle control.

For offline meta-learning, we collected a dataset comprised of approximately $1,700,000$ trajectories ($9.5$ hours) of autonomous driving that includes adverse weather conditions like rain.
Approximately $60\%$ is from the Mojave Desert near Helendale, CA; $30\%$ from Halter Ranch near Paso Robles, CA; $10\%$ from coastal sage near Oceanside, CA; and $5\%$ from coastal dunes also nearby Oceanside, CA.
At these locations, we collect data from a diverse range of terrains that are depicted and described in \cref{fig:varied_terrains}.
We also provide statistics on key dataset quantities in \cref{table:dataset}.

For each trajectory, we use an adaptation length of $\adaptlen = 1,000$ steps (\SI{20}{\second}) and a prediction length of $\mppiwindow = 250$ steps (\SI{5}{\second}), with discrete time steps spaced (\SI{0.02}{\second}) apart.
Training follows \cref{alg:meta} for 
$\numepochs=20$ epochs, starting with a 5-epoch pretraining phase without meta-learning (disabling adaptation by setting  $\adaptparams_t = 0$), followed by 15 epochs with meta-learning enabled.
For comparison, a baseline model was also trained for 20 epochs on the exact same dataset without meta-learning.

We perform online adaptation (\cref{alg:adapt}) at a rate of \SI{5}{\hertz} by updating every $\numsteps=10$ time steps.
For control, \ac{mppi} (\cref{sec:ctrl}) runs at \SI{30}{\hertz}, leveraging the most recent set of adapted parameters $\adaptparams$ for forward predictions.
The car completes 3 laps around a figure-8 track with additional curves added to increase planning difficulty (see \cref{fig:front}).
We test the {\baseline} and {\ours} configurations for 4 repeated runs each.

To evaluate \cref{hypoth:pred}, we calculate the average model prediction error over each run, defined as the Euclidean distance between the endpoints of the predicted and actual trajectories.
For performance assessment (\cref{hypoth:perf}), we log the completion time and average speed of each run.
Safety (\cref{hypoth:perf}) is analyzed by tracking the number of instances the car exceeds safety limits, the duration spent outside these limits, and the associated cost for both track boundary and rollover ratio limit violations.

\subsubsection{Results}
\Cref{fig:allruns} compares trajectories for the {\baseline} and {\ours} configurations and \cref{fig:front} shows video stills capturing the adaptation configuration in action.
The {\baseline} configuration results in erratic behavior, with frequent violations of track boundaries and rollover constraints.
In contrast, the {\ours} configuration yields more stable and compliant motion, as the dynamics model is adapted in real-time (\cref{fig:thetanorm}).

The quantitative results in \cref{tab:hardware} further support the advantages of adaptation.
The adapted model achieves significantly lower prediction error, confirming that online adaptation effectively learns the terrain dynamics, validating \cref{hypoth:pred}.
Importantly, the {\ours} configuration not only improves performance—achieving faster track completion times and higher average speeds—but also enhances safety metrics.
Specifically, the adapted vehicle crosses safety limits (e.g., track boundaries and rollover thresholds) far less frequently and spends significantly less time in unsafe states compared to the baseline.
These improvements translate to a dramatic reduction in associated track and rollover costs, highlighting the effectiveness of the adaptation method, confirming \cref{hypoth:perf}.

In high-speed off-road driving, an inaccurate dynamics model can lead MPPI to generate unsafe trajectories that fail to account for real-world dynamics.
By incorporating accurate, environment-adapted predictions, our adaptation method enables MPPI to generate control sequences that balance performance and safety.
These results underscore the importance of real-time model adaptation in ensuring both reliable and secure autonomous vehicle operation in challenging terrains.

\begin{table*}[th]
    \centering
    \caption{
    \footnotesize{
    Results of the simulated experiments.
    The means over 25 runs of each configuration are displayed.
    }
    }
    \begin{tabular}{lcccccccccccccc}
    \hline
        ~ & \multicolumn{4}{c}{Completion time, \si{\second}} & ~ &\multicolumn{4}{c}{Average speed, \si{\meter\per\second}} & ~ & \multicolumn{4}{c}{Prediction Error, \si{\meter}} \\ \hline
        Steepness & shallow & shallow & steep & steep & ~ & shallow & shallow & steep & steep & ~ & shallow & shallow & steep & steep \\ 
        Obstacle Density & sparse & dense & sparse & dense & ~ & sparse & dense & sparse & dense & ~ & sparse & dense & sparse & dense \\ \hline 
        \baseline & \textbf{39.2} & 44.4 & \textbf{73.0}\confint & \textbf{86.2}\confint & ~ & \textbf{5.88}\confint & 5.32 & \textbf{4.06}\confint & \textbf{3.59}\confint & ~ & 7.19 & 7.62 & 8.17 & 9.03 \\ 
        {\leastsq} & {46.7} & {46.6} & {87.7} & {102.0} & ~ & {5.04} & {5.06} & {3.43} & {3.06} & ~ & {5.71} & {5.00} & {5.21} & {5.35}
        \\
        \nometa & 40.4 & \textbf{43.1}\confint & 83.0 & 95.3 & ~ & 5.70 & \textbf{5.47}\confint & 3.59 & 3.15 & ~ & 4.54 & 4.45 & 4.25 & \textbf{4.64} \\ 
        \ours & 40.1 & 44.7 & 88.8 & 105.9 & ~ & 5.75 & 5.27 & 3.34 & 2.99 & ~ & \textbf{2.96}\confint & \textbf{2.19}\confint & \textbf{3.67}\confint & 4.65 \\ \hline
        ~ & ~ & ~ & ~ & ~ & ~ & ~ & ~ & ~ & ~ & ~ & ~ & ~ \\ 
        ~ & \multicolumn{4}{c}{\# times over rollover limit} & ~ & \multicolumn{4}{c}{Time exceeding rollover limit, \si{\second}} & ~ & \multicolumn{4}{c}{Rollover Cost ($\times 10^4$)} \\ \hline
        \baseline & 2.7 & \textbf{3.6} & 5.4 & 9.9 & ~ & 1.30 & 5.31 & 5.09 & 7.50 & ~ & 1.09 & 1.75 & 1.82 & 8.91 \\ 
        {\leastsq} & {2.5} & {4.1} & {5.4} & {7.1} & ~ & {2.19} & {6.12} & {4.24} & {5.53} & ~ & {1.19} & {2.00} & {4.23} & {8.59}
        \\
        \nometa & 2.9 & 4.6 & 4.7 & 6.7 & ~ & 1.47 & 7.10 & 5.71 & 5.42 & ~ & 1.04 & 2.49 & 1.14 & 2.75 \\ 
        \ours & \textbf{1.9} & 5.3 & \textbf{3.2}\confint & \textbf{6.2} & ~ & \textbf{0.97} & \textbf{4.83} & \textbf{3.88} & \textbf{3.40}\confint & ~ & \textbf{0.99} & \textbf{1.15}\confint & \textbf{0.83} & \textbf{2.57} \\ \hline
        \multicolumn{15}{l}{\confint Significant best result: bootstrapped 95\% confidence intervals of the mean do not overlap with those of any other configuration.}
    \end{tabular}
    \label{tab:sim}
\end{table*}

\subsection{Simulated Experiments}
We perform extensive simulated experiments to further confirm our hypotheses.

\subsubsection{Experimental Setup}
The simulated experiments use the same baseline and meta-learned models trained for the real-world validation (\cref{sec:realexp}) and identical adaptation parameters.
However, the physics simulator employs distinct dynamics  based on the bicycle model, creating a real2sim transfer scenario where adaptation must account for the dynamics mismatch between the real-world training data and the simulated environment.
\begin{figure}[tb]
    \centering
    \includegraphics[width=.48\textwidth]{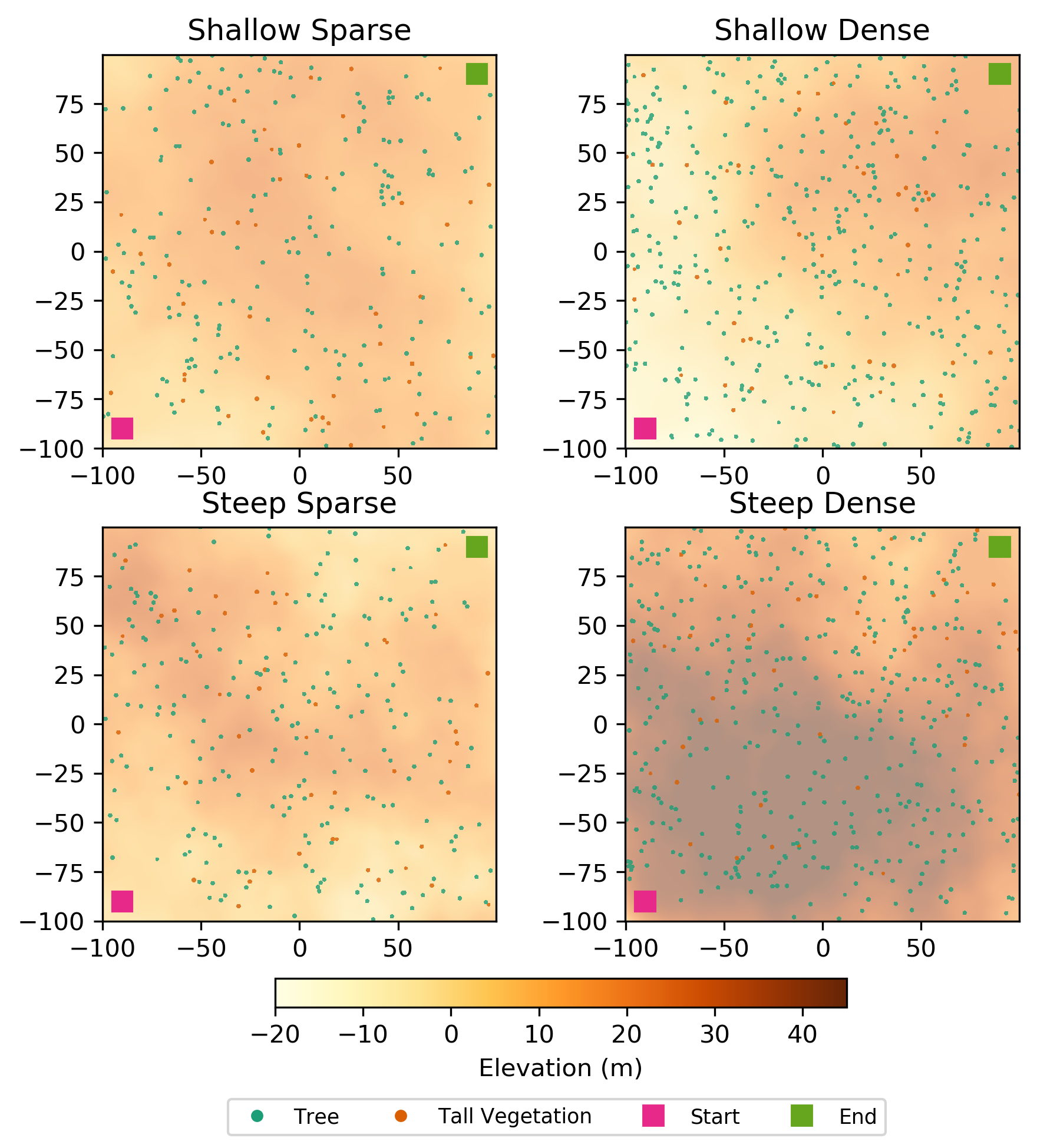}
    \caption{\footnotesize{
    Procedurally generated maps for the simulated experiments; horizontal and vertical axes are in meters.
    }}
    \label{fig:terrain}
\end{figure}
We procedurally generate four diverse maps (\cref{fig:terrain}), each with unique terrain and obstacle configurations.
The maps are categorized as \emph{shallow-sparse}, \emph{shallow-dense}, \emph{steep-sparse}, and \emph{steep-dense}, where the first term denotes the steepness of terrain features, and the second describes the density of obstacles.
This variety ensures that the simulated environment tests the adaptation across a wide range of scenarios.

Everything else remains consistent with the real-world validation, except for a few key modifications.
First, since the maps are open with sufficiently distant boundaries, track costs are omitted from the analysis.
Next, we evaluate two new configurations: {\nometa} and {\leastsq}. 
In the {\nometa} configuration, online adaptation is applied to the \emph{baseline} model to assess the impact of not meta-learning the basis functions and Kalman filter parameters.
For the {\leastsq} configuration, we implement the method of adaptation from \cite{nagy2023ensemble}, which consists of using sliding-window regularized least squares to adapt the weights of an ensemble model in real time.
To ensure statistical significance, we perform 25 runs per configuration on each map.

\subsubsection{Results}
The results of the simulated experiments are summarized in \cref{tab:sim}.
As expected, prediction errors in simulation are generally higher than in real-world testing due to the real2sim gap.
Nevertheless, all adaptation configurations outperform the baseline in prediction accuracy, with the meta-learned adaptation mostly achieving the lowest prediction errors, further confirming \cref{hypoth:pred}.

In terms of performance, the baseline configuration generally achieves shorter completion times and higher average speeds.
However, this comes at the cost of compromised safety, with the baseline configuration exhibiting higher rollover occurrences, longer time spent exceeding limits, and increased safety costs.
In contrast, the adaptation configurations demonstrate significantly improved safety, with the meta-learned configuration generally achieving the greatest reductions across all safety metrics.
These findings align with the real-world validation results, further supporting \cref{hypoth:perf}.

Crucially, the comparison between our meta-learned configuration and the non-meta-learned adaptation configurations, {\nometa} and {\leastsq}, highlights the importance of meta-learning.
Our configuration generally outperforms in prediction error and safety metrics across all map types.
The non-meta-learned configurations rely on suboptimally selected basis functions, and the {\leastsq} configuration requires hand tuning of the window length and regularization parameter---directly influencing the adaptation rate.
In contrast, our method learns both the basis functions and adaptation dynamics from data, resulting in models that adapt more effectively, yield better closed-loop performance, and bridge the substantial dynamics mismatch from the real2sim gap.

\section{Conclusion} 
\label{sec:conclusion}

In this work, we introduced a meta-learning framework for online dynamics model adaptation applied to high-speed off-road autonomous driving.
By combining a Kalman filter-based adaptation scheme with meta-learned parameters, our approach addresses the challenges of unseen or evolving terrain dynamics, enhancing prediction accuracy, performance, and safety.
Empirical validation through real-world and simulated experiments demonstrates that our method outperforms baseline and non-meta-learned adaptation strategies, particularly in safety-critical scenarios.
Our method is applicable to a broad range of model-based control scenarios beyond off-road autonomous driving, as it can be integrated with any model that is linear in its adaptable parameters and any \ac{mpc}-type controller. 
This contribution represents a step toward more robust and reliable autonomous systems capable of adapting to complex and changing environments.

\subsection{Limitations}
While our approach demonstrates significant improvements in prediction accuracy and safety, several limitations remain.
The initial selection of Kalman filter parameters is still user-defined, which could lead to unstable training if poorly chosen.
Another key challenge is the method's reliance on sufficiently exciting inputs to effectively identify the adapted parameters.
For instance, driving straight on flat terrain may not produce enough variability for meaningful adaptation. Future work could explore active exploration strategies to ensure effective parameter identification.
Finally, the computational overhead during meta-learning is another consideration, as taking gradients through the adaptation process increases memory usage and training time.
To address this, future work could investigate techniques to limit gradient propagation through only a subset of parameters, such as $\params^{l}$, reducing resource requirements.

\section*{Acknowledgments}
The authors would like to thank Trey Smith and Brian Coltin for their helpful insights and feedback. The research was carried out at the Jet Propulsion Laboratory, California Institute of Technology, under a contract with the National Aeronautics and Space Administration (80NM0018D0004). This work was supported by a NASA Space Technology Graduate Research Opportunity under award 80NSSC23K1192, by the National Science Foundation under Grant No. 2409535, and by Defense Advanced Research Projects Agency (DARPA). Approved for Public Release, Distribution Unlimited. ©2025. All rights reserved.


\bibliographystyle{plainnat}
\bibliography{references}

\end{document}